\documentclass[twoside]{article}
\usepackage[utf8]{inputenc}
\usepackage{amsmath}
\usepackage{amsfonts}
\usepackage{amssymb}
\usepackage{graphicx}
\usepackage{color}

\usepackage[svgnames]{xcolor}
\usepackage{xcolor,colortbl} 

\usepackage[normalem]{ulem}
\usepackage[left=2cm,right=2cm,top=2cm,bottom=2cm]{geometry} 

\newtheorem{theorem}{Theorem}[section]

\newcommand{\qed}{\hfill$\Box$\par\medskip}

\DeclareMathOperator{\R}{\mathbb{R}}
\def\bhag#1{\noindent
\setcounter{equation}{0}
\section{#1}
}

\def\RR{{\mathbb R}}

\def\ZZ{{\mathbb Z}}

\def\PPI{{{\rm I}\kern-1pt\Pi}}
\def\SS{{\mathbb S}}

\def\b #1;{{\bf #1}}
\def\x{{\bf x}}
\def\k{{\bf k}}
\def\y{{\bf y}}
\def\u{\mathbf{u}}
\def\w{{\bf w}}
\def\z{{\bf z}}

\def\v{\mathbf{v}}

\def\j{\mathbf{j}}

\def\e{\epsilon}

\def\O{{\cal O}}

\def\C{{\mathcal C}}

\def\ip#1#2{{\langle {#1}, {#2}\rangle}}

\def\esssup{\mathop{\hbox{{\rm ess sup}}}}

\def\be{\begin{equation}}
\def\ee{\end{equation}}
\def\bea{\begin{eqnarray}}
\def\eea{\end{eqnarray}}
\def\eref#1{(\ref{#1})}
\def\disp{\displaystyle}

\def\binom#1#2{\small{\left(\!\!\begin{array}{c}{#1}\\{#2}
\end{array}\!\!\right)}}
\def\donchitre#1#2{\vskip 6.5cm\noindent
\parbox[t]{1in}{\special{eps:#1.eps x=6.5cm y=5.5cm}}
\hbox to 7cm{}\parbox[t]{0.0cm}{\special{eps:#2.eps x=6.5cm y=5.5cm}}}

\def\XX{{\mathbb X}}

\def\corr#1{#1}

\newcommand*{\titleAT}{\begingroup
  \newlength{\drop}
  \drop=0.05\textheight
  \begin{center}
  \includegraphics[scale=0.4]{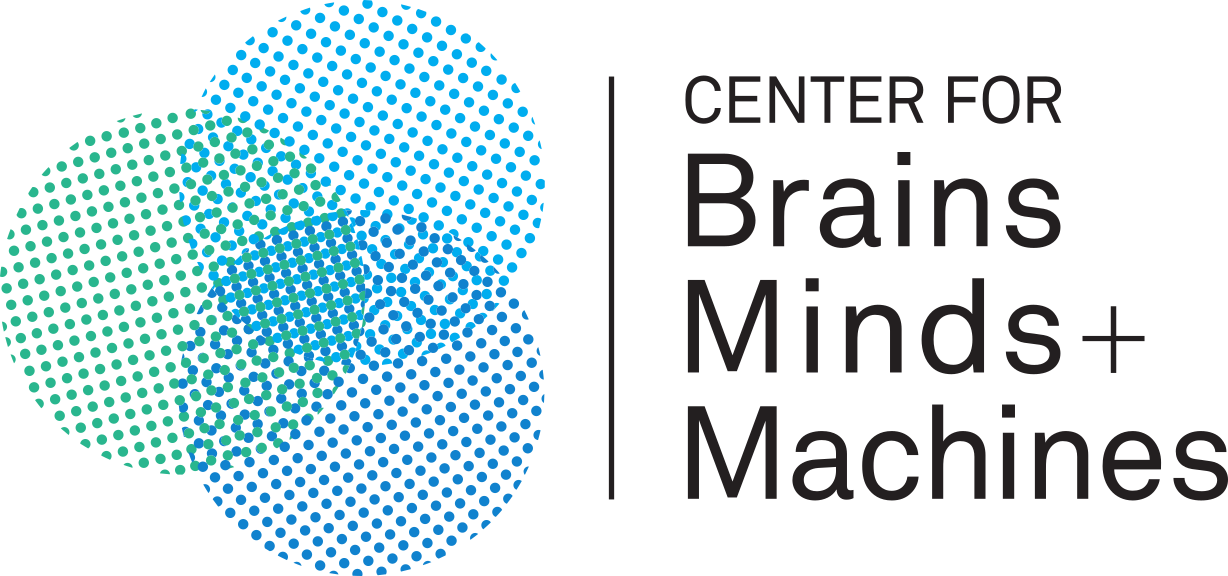} 
  \end{center} 
  \vspace{2pt}\vspace{-\baselineskip}

  \vspace{\drop}
  \textbf{\large{CBMM Memo No. \memonumber}}   \hfill    \textbf{\large{\memodate}} 

  \vspace{\drop}
  \begin{center}
    \textbf{\huge{\memotitle}}\\
    \vspace{0.4\drop}
    \textbf{\Large{by}}\\
    \vspace{0.4\drop}
    \large{\memoauthors}
  \end{center}
  \vspace{\drop}
  \textbf{\large{\noindent Abstract}:} {\memoabstract}

\vspace{\fill}
  \rule{\textwidth}{0.4pt}\par

  \begin{minipage}{.15\linewidth}
    \includegraphics[scale=0.1]{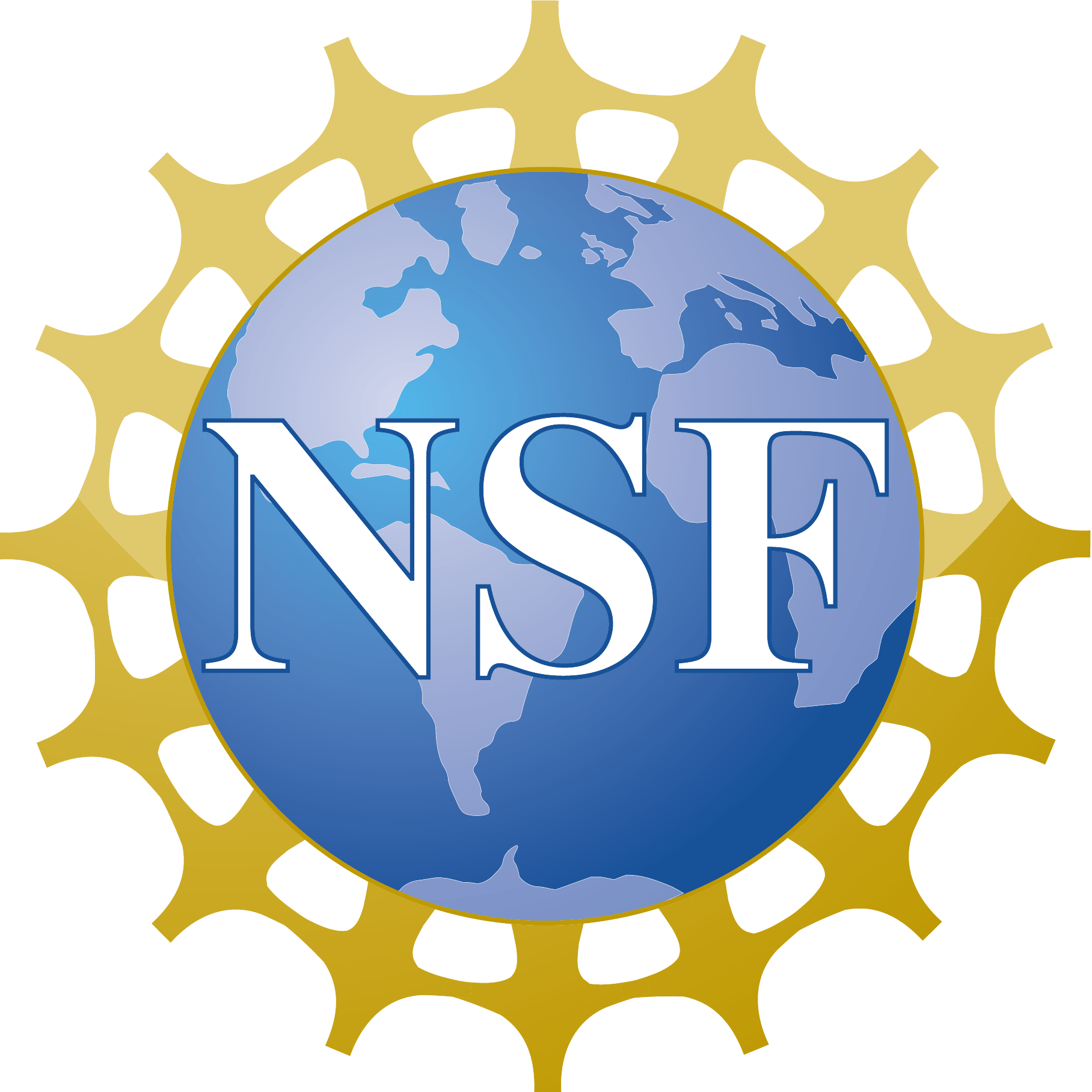}
  \end{minipage}
  \begin{minipage}{.84\linewidth}
    \textbf{\large{This work was supported by the Center for Brains, Minds and Machines (CBMM), funded by NSF STC award  CCF - 1231216.
        H.M. is supported in part by ARO Grant W911NF-15-1-0385.}}
  \end{minipage}
  \endgroup}

\begin{document}

\def\memonumber{054}
\def\memodate{\today}
\def\memotitle{Deep vs. Shallow Networks: an Approximation Theory Perspective}
\def\memoauthors{ \textbf{ Hrushikesh N. Mhaskar$^1$ and   Tomaso Poggio$^2$ \\
\small
1. Department of Mathematics, California Institute of Technology, Pasadena, CA 91125 \\
   Institute of Mathematical Sciences, Claremont Graduate University, Claremont, CA 91711. \\
\textsf{hrushikesh.mhaskar@cgu.edu} \\
2. Center for Brains, Minds, and Machines, McGovern Institute for Brain Research, \\
   Massachusetts Institute of Technology, Cambridge, MA, 02139.  \\
   \textsf{tp@mit.edu} \\
  }
}

\normalsize
\def\memoabstract{
The paper briefly reviews several recent results on hierarchical
    architectures for learning from examples, that may formally
    explain the conditions under which Deep Convolutional Neural
    Networks perform much better in function approximation problems
    than shallow, one-hidden layer architectures. The paper announces
    new results for a non-smooth activation function -- the ReLU
    function -- used in present-day neural networks, as well as for
    the Gaussian networks. We propose a new definition of {\it
      relative dimension} to encapsulate different notions of sparsity
    of a function class that can possibly be exploited by deep
    networks but not by shallow ones to drastically reduce the
    complexity required for approximation and learning.}

\titleAT

\newpage


\bhag{Introduction}\label{intsect}

Deep Neural Networks especially of the convolutional type (DCNNs) have
started a revolution in the field of artificial intelligence and
machine learning, triggering a large number of commercial ventures and
practical applications. Most deep learning references
these days start with Hinton's backpropagation and with Lecun's
convolutional networks (see for a nice review
\cite{lecun2015deep}). Of course, multilayer convolutional
networks have been around at least as far back as the optical
processing era of the 70s. Fukushima's
Neocognitron \cite{fukushima1980} was a convolutional neural network
  that was trained to recognize characters. The HMAX model of visual
  cortex \cite{Riesenhuber1999} was described as a series of AND and
  OR layers to represent hierarchies of disjunctions of conjunctions.
  A version of the questions about the importance of hierarchies was asked in
  \cite{poggio03mathematics} as follows: ``\textit{A comparison with real
    brains offers another, and probably related, challenge to learning
    theory. The ``learning algorithms'' we have described in this
    paper correspond to one-layer architectures. Are hierarchical
    architectures with more layers justifiable in terms of learning
    theory? It seems that the learning theory of the type we have
    outlined does not offer any general argument in favor of
    hierarchical learning machines for regression or classification.
    This is somewhat of a puzzle since the organization of cortex --
    for instance visual cortex -- is strongly hierarchical.  At the
    same time, hierarchical learning systems show superior performance
    in several engineering applications.}''

   Ironically a mathematical theory
characterizing the  properties of DCNN's and even simply why they work so well
is still missing. 
Two of the basic theoretical questions about Deep Convolutional Neural Networks
(DCNNs) are:
\begin{itemize}
\item which classes of functions can  they approximate well?
\item why is stochastic gradient descent (SGD) so unreasonably efficient?
\end{itemize}

In this paper we review and extend a theoretical framework that we
have introduced very recently to address the first question
\cite{mhaskar_poggio_uai_2016}. The theoretical results include
answers to why and when deep networks are better than shallow by using
the idealized model of a deep network as a directed acyclic graph
(DAG), which we have shown to capture the properties a range of
convolutional architectures recently used, such as the very deep
convolutional networks of the ResNet type
\cite{EXTREMELYDEEP_MS_2015}. For compositional functions conforming
to a DAG structure with a small maximal indegree of the nodes, such
as a binary tree structure, one can bypass the curse of
dimensionality with the help of the blessings of compositionality
(cf. \cite{donoho2000high} for a motivation for this terminology).  We demonstrate this fact using three
examples : traditional sigmoidal networks, the ReLU networks commonly
used in DCNN's, and Gaussian networks.  The results announced for the
ReLU and Gaussian networks are new.  We then give examples of
different notions of sparsity for which we expect better performance
of DCNN's over shallow networks, and propose a quantitative
measurement, called relative dimension, encapsulating each of these
notions, independently of the different roles the various parameters
play in each case.

In Section~\ref{compfuncsect}, we explain the motivation for
considering compositional functions, and demonstrate how some older
results on sigmoidal networks apply for approximation of these
functions.  In Section~\ref{shallowsect}, we announce our new results
in the case of shallow networks implementing the ReLU and Gaussian
activation functions.  The notion of a compositional function
conforming to a DAG structure is explained in Section~\ref{deepsect},
in which we also demonstrate how the results in
Section~\ref{shallowsect} lead to better approximation bounds for such
functions.  The ideas behind the proofs of these new theorems are
sketched in Section~\ref{ideasect}.  Finally, we make some concluding
remarks in Section~\ref{sparssect}, pointing out a quantitative
measurement for three notions of sparsity which we feel may be
underlying the superior performance of deep networks.

\bhag{Compositional functions}\label{compfuncsect}
The purpose of this section is to introduce the concept of compositional functions, and illustrate by an example how this leads to a better approximation power for deep networks. 
In Sub-section~\ref{motivsect}, we explain how such functions arise in image processing and vision. 
In Sub-section~\ref{reviewsect}, we review some older results for approximation by shallow networks implementing a sigmoidal activation function, and explain how a ``good error propagation'' helps to generalize these results for deep networks.

\subsection{Motivation}\label{motivsect}

Many of the computations performed on images should reflect the
symmetries in the physical world that manifest themselves through the
image statistics. Assume for instance that a computational hierarchy
such as \be\label{compositionfigure}
h_l(\cdots h_3(h_{21} (h_{11}(x_1, x_2), h_{12}(x_3, x_4)),\\
h_{22}(h_{13}(x_5, x_6), h_{14}(x_7, x_8))\cdots))) \ee is given. Then
shift invariance of the image statistics is reflected in the following
property: the local node ``processors'' satisfy $h_{21}=h_{22}$ and
$h_{11}=h_{12}=h_{13}=h_{14}$ since there is no reason for them to be
different across an image. Similar invariances of image statistics --
for instance to scale rotation -- can be similarly used to constrain
visual algorithms and their parts such as the local processes $h$.

It is natural to ask whether the hierarchy itself -- for simplicity
the idealized binary tree of the Figure \ref{example_binary} --
follows from a specific symmetry in the world and which one.  A
possible answer to this question follows from the fact that in natural
images the target object is usually among several other objects at a
range of scales and position. From the physical point of view, this is
equivalent to the observation that there are several localized
clusters of surfaces with similar properties (object parts, objects,
scenes, etc). These basic aspects of the physical world are reflected
in properties of the statistics of images: {\it locality, shift
  invariance and scale invariance}. In particular, locality reflects
clustering of similar surfaces in the world -- the closer to each
other pixels are in the image, the more likely they are to be
correlated. Thus nearby patches are likely to be correlated (because
of locality), at all scales.  Ruderman's pioneering work
\cite{Ruderman1997} concludes that this set of properties is {\it
  equivalent to the statement that natural images consist of many
  object patches that may partly occlude each other} (object patches
are image patches which have similar properties because they are
induced by local groups of surfaces with similar properties). We argue
that Ruderman's conclusion reflects the compositionality of objects
and parts: parts are themselves objects, that is self-similar clusters
of similar surfaces in the physical world. The property of {\it
  compositionality} was in fact a main motivation for hierarchical
architectures such as Fukushima's and later imitations of it such as
HMAX which was described as a pyramid of AND and OR layers
\cite{Riesenhuber1999}, that is a sequence of conjunctions and
disjunctions. According to these arguments, compositional functions
should be important for vision tasks because they reflect constraints
on visual algorithms. 

The following argument shows that compositionality of visual
computations is a basic property that follows from the simple
requirement of {\it scalability} of visual algorithms: an algorithm
should not change  if the size of the image (in pixels) changes. In
other words, it should be possible to add or subtract simple reusable
parts to the algorithm to adapt it to increased or decreased size of
the image without changing its basic core. 

A way to formalize the argument is the following. Consider the class
of nonlinear functions, mapping vectors from $\RR^n$
into $\RR^d$ (for simplicity we put in the following $d=1$). Informally
we call an algorithm $K_n: \RR^n \mapsto \RR$ {\it scalable} if it maintains
the same ``form'' when the input vectors increase in dimensionality;
that is, the same kind of computation takes place when the size of the
input vector changes. Specific definitions of scalability and shift
invariance for any (one-dimensional) image size lead to the following
characterization
of scalable, shift-invariant functions or algorithms:
\textit{ Scalable, shift-invariant functions $K: \RR^{2m}
  \mapsto \RR$ have the structure $K = H_2 \circ H_4 \circ H_6 \cdots
  \circ H_{2m}$, with $H_4=\tilde{H}_2 \oplus \tilde{H}_2$, $H_6=H^*_2 \oplus H^*_2 \oplus
  H^*_2$, etc., \corr{where $\tilde{H}_2$ and $H^*_2$ are suitable functions}}.

\begin{figure}\centering
\includegraphics[width=0.5\textwidth]{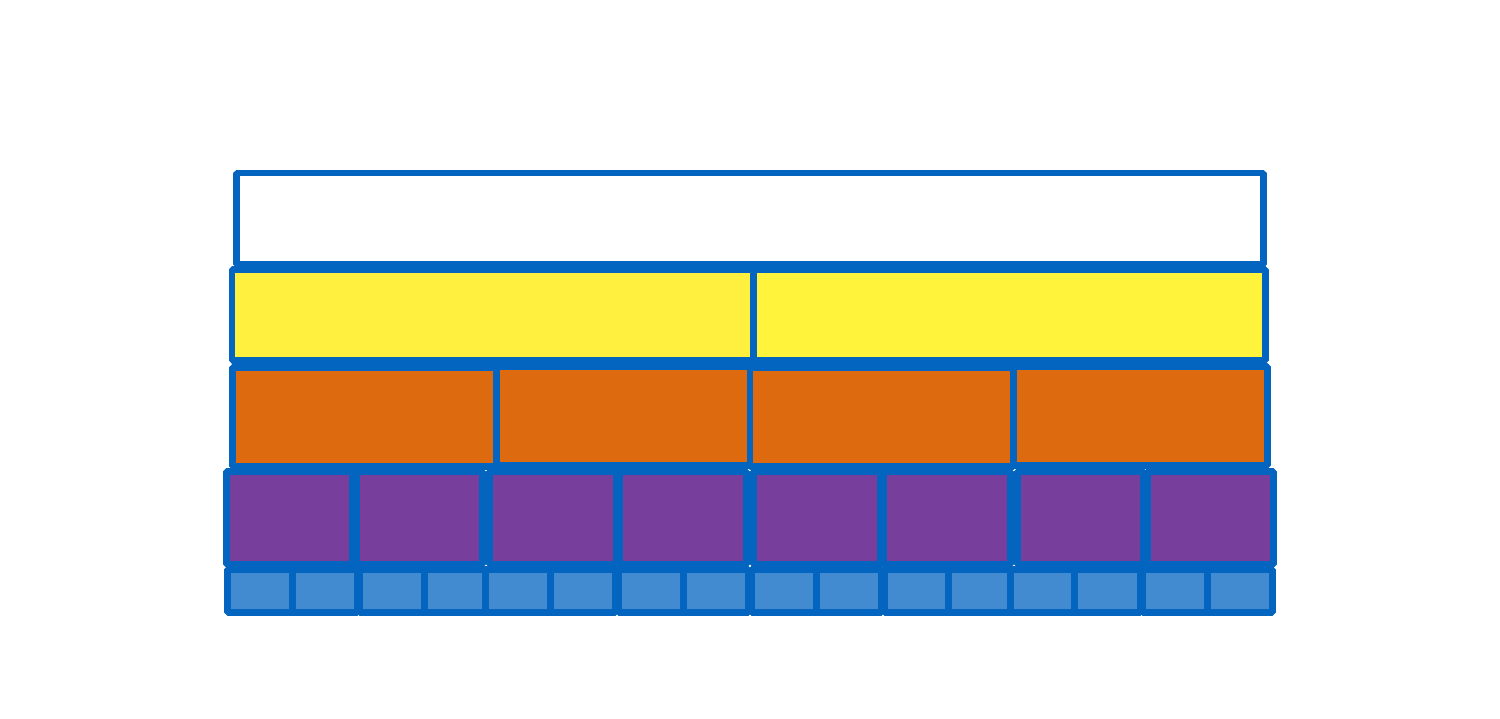}
\caption{\it A scalable function.  Each
layer consists of identical blocks; each block is a function $H_{2}: \RR^2
\mapsto \RR$. The overall function shown in the figure is $ \R^{32}
\mapsto \RR$ }
\label{ScalableOperator} 
\end{figure}

Thus the structure of {\it shift-invariant, scalable functions}
consists of several layers; each layer consists of identical blocks;
each block is a function $H: \RR^2 \mapsto \RR$: see Figure~\ref{ScalableOperator}. Obviously, shift-invariant scalable functions are
equivalent to shift-invariant compositional functions. The
definition can be changed easily in several of its specifics. For instance
for two-dimensional images the blocks could be operators $H: \RR^5
\rightarrow \RR$ mapping a neighborhood around each pixel into a real
number.

The final step in the argument uses the universal approximation property
to claim that a nonlinear node with two inputs and enough units (that
is, channels) can approximate arbitrarily well each of the $H_2$
blocks. This leads to conclude that deep convolutional neural networks are natural
approximators of {\it scalable, shift-invariant functions}.

\subsection{An example}\label{reviewsect}

In this section, we illustrate the advantage of approximating a compositional function using deep networks corresponding to the compositional structure rather than a shallow network that does not take into account this structure.

In the sequel, for any integer $q\ge 1$, $\x=(x_1,\cdots,x_q)\in\RR^q$, $|\x|$ denotes the Euclidean $\ell^2$ norm of $\x$, and $\x\cdot\y$ denotes the usual inner product between $\x,\y\in\RR^q$. In general, we will not complicate the notation by mentioning the dependence on the dimension in these notations unless this might lead to confusion.

  Let
$I^q=[-1,1]^q$, $\XX=C(I^q)$ be the space of all continuous functions
on $I^q$, with $\|f\|=\max_{\x\in I^q}|f(\x)|$. If $\mathbb{V}\subset\XX$, we define $\mathsf{dist}(f,\mathbb{V})=\inf_{P\in\mathbb{V}}\|f-P\|$. Let $\mathcal{S}_n$
denote the class of all shallow networks with $n$ units of the form
$$
\x\mapsto\sum_{k=1}^n a_k\sigma(\mathbf{w}_k\cdot\x+b_k),
$$
where $\mathbf{w}_k\in\RR^q$, $b_k, a_k\in\RR$. The number of trainable
parameters here is $(q+2)n\sim n$. Let $r\ge 1$ be an integer, and
$W_{r,q}^{\mbox{NN}}$ be the set of all functions with continuous partial
derivatives of orders up to $r$ such  that $\|f\|+\sum_{1\le |\k|_1\le r} \|D^\k f\| \le 1$, where $D^\k$ denotes the partial derivative indicated by the
multi--integer $\k\ge 1$, and $|\k|_1$ is the sum of the components of
$\k$.

\begin{figure}[h]
\centering
\includegraphics[width=0.9\textwidth]{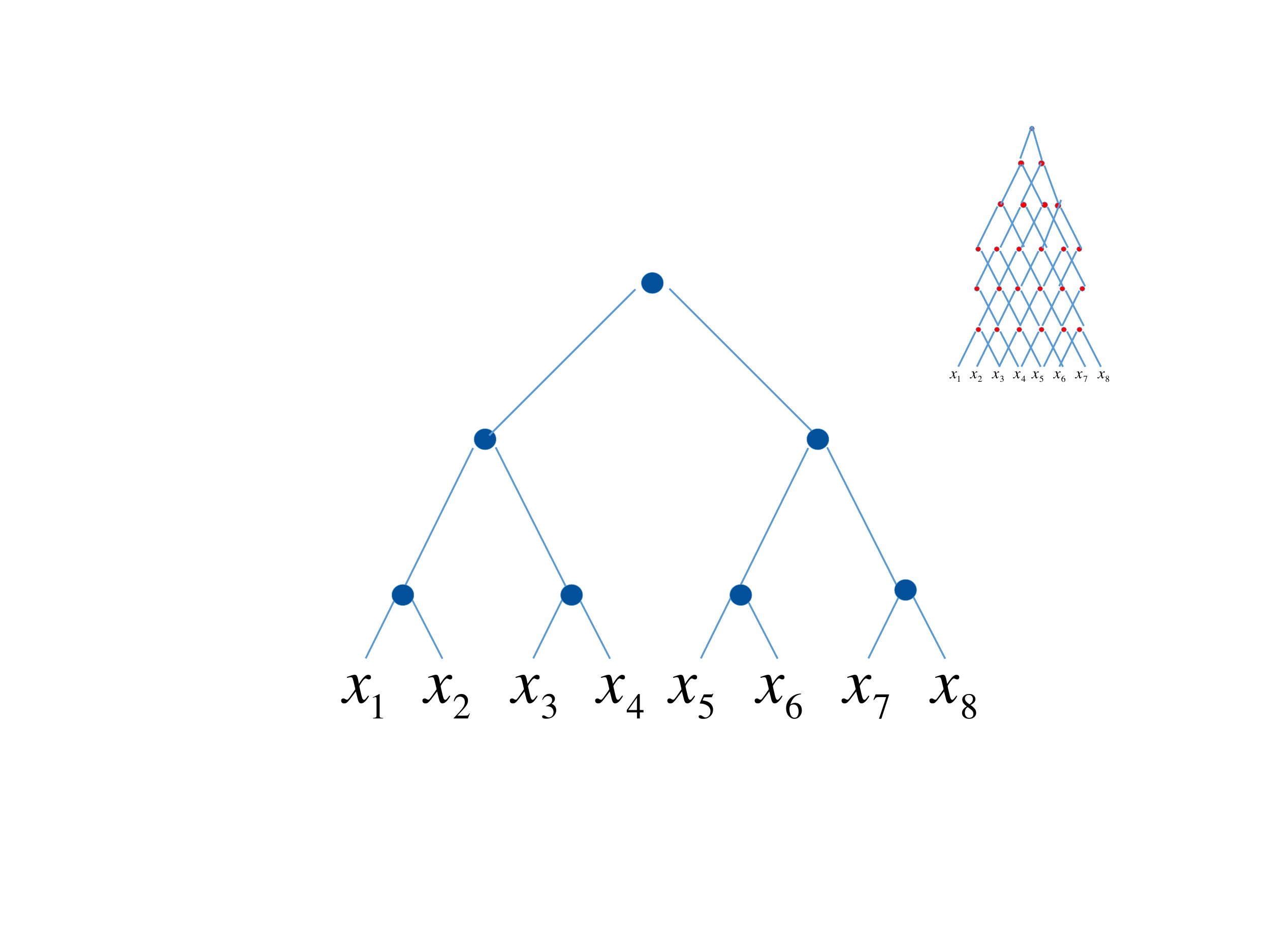}
\caption{\it  A  binary tree hierarchical network in 8 variables, which approximates
 well functions of the form \eref{l-variables}.  Each of the nodes consists of $n$ 
  units and computes the ridge function   
  $\sum_{i=1}^n a_i\sigma(\ip{\mathbf{v}_i}{\mathbf{x}}+t_i)$, with
  $\mathbf{v}_i, \mathbf{x} \in \RR^2$, $a_i, t_i\in\RR$. 
 Similar to the shallow network such a
  hierarchical network can approximate any continuous function;
  the text proves how it approximates  compositional functions better
  than a shallow network.  Shift invariance may additionally hold
  implying that the weights in each layer are the same. The inset at
  the top right shows a network similar to ResNets: our results on
  binary trees apply to this case as well with obvious changes in the constants}
\label{example_binary}
\end{figure}

For explaining our ideas for the deep network, we consider compositional functions conforming to a binary tree. For example, we consider functions of the form (cf. Figure~\ref{example_binary})

\be\label{l-variables}
f(x_1,
  \cdots, x_8) = h_3(h_{21}(h_{11} (x_1, x_2),  h_{12}(x_3, x_4)),
  h_{22}(h_{13}(x_5, x_6),  h_{14}(x_7, x_8))).
  \ee

\begin{figure}[h]
\centering
\includegraphics[width=0.45\textwidth,  height=0.3\textwidth]{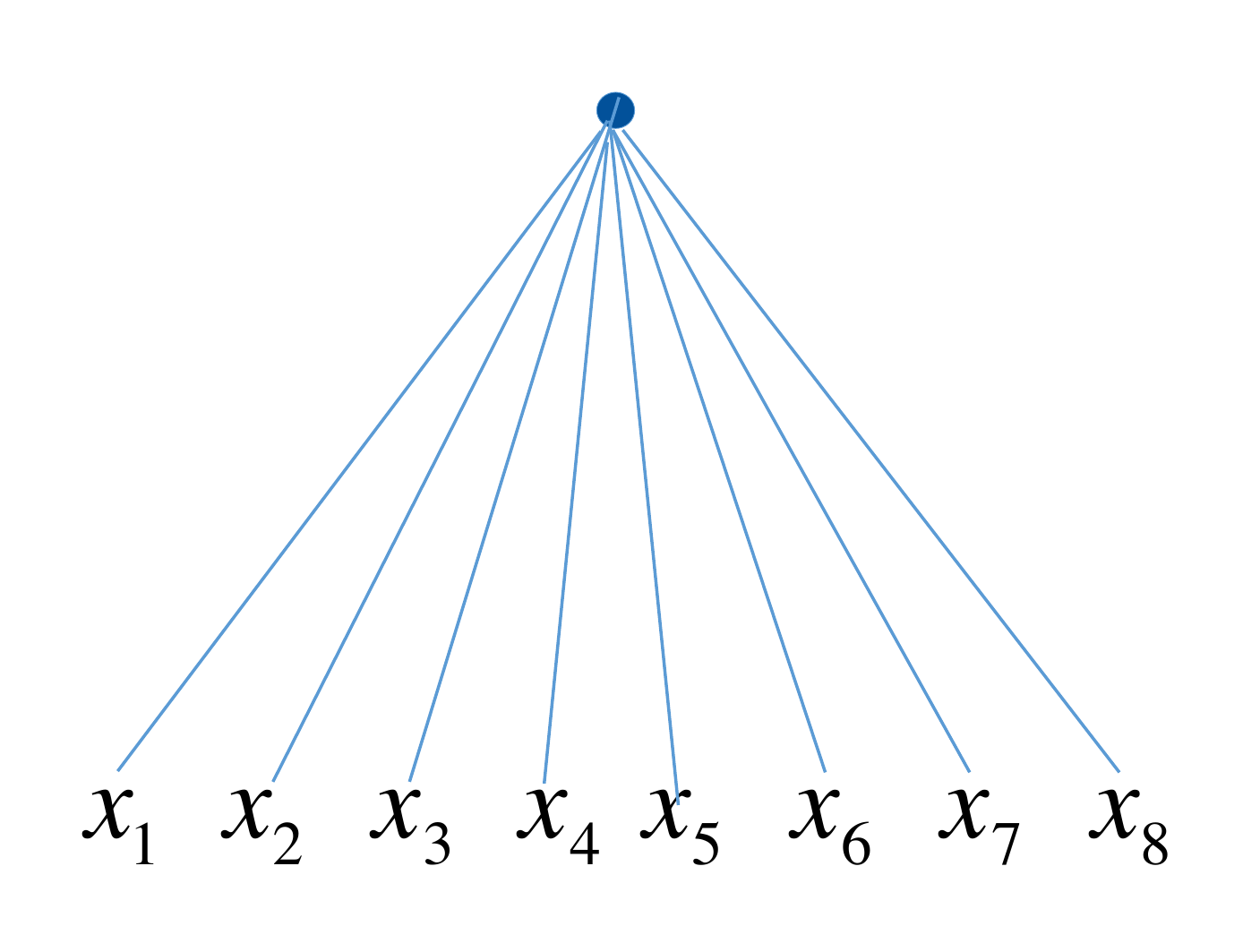}  
\caption{\it A shallow universal network in 8 variables and $N$ units
  which can approximate a generic function $f(x_1, \cdots, x_8)$. 
 The top node consists of $n$ 
  units and computes the ridge function   
  $\sum_{i=1}^n a_i\sigma(\ip{\mathbf{v}_i}{\mathbf{x}}+t_i)$, with
  $\mathbf{v}_i, \mathbf{x} \in \RR^2$, $a_i, t_i\in\RR$.}
\label{example_shallow}
\end{figure}

For the hierarchical binary tree network, the  spaces analogous to $W_{r,q}^{\mbox{NN}}$ are  
 $W_{H,r,2}^{\mbox{NN}}$,  defined to be the
class of all functions $f$ which have the same structure (e.g.,
(\ref{l-variables})), where each of the constituent functions $h$ is
in $W_{r,2}^{\mbox{NN}}$ (applied with only $2$ variables).  We define the corresponding
class of deep networks $\mathcal{D}_n$ to be set of all functions with
the same structure, where each of the constituent functions  is in
$\mathcal{S}_n$. We note that in the case when $q$ is an integer power of $2$, the number of parameters involved in an element of
$\mathcal{D}_{n}$ -- that is, weights and biases, in a node of the binary tree  is $(q-1)(q+2)n$.

The following theorem   (cf. \cite{optneur}) estimates the degree of approximation for shallow and deep networks. We remark that the assumptions on $\sigma$ in the theorem below are not satisfied by the
ReLU function $x\mapsto |x|$, but they are satisfied by smoothing the function in an
arbitrarily small interval around the origin. 

\begin{theorem}
\label{optneurtheo}
Let $\sigma :\RR\to \RR$ be infinitely differentiable, and not a polynomial on any subinterval of $\RR$. \\
{\rm (a)} For $f\in W_{r,q}^{\mbox{NN}}$
\be\label{optneurest}
\mathsf{dist}(f,\mathcal{S}_n)= \O(n^{-r/q}).
\ee 
{\rm (b)} For $f\in W_{H,r,2}^{\mbox{NN}}$
\begin{equation}
\mathsf{dist}(f,\mathcal{D}_n) =\mathcal{O}(n^{-r/2}).
\label{deepnetapprox}
\end{equation}
\end{theorem}

\noindent\textit{Proof.}
Theorem~\ref{optneurtheo}(a) was proved by
  \cite{optneur}. To prove Theorem~\ref{optneurtheo}(b), we observe that
each of the constituent functions being in $W_{r,2}^{\mbox{NN}}$,
(\ref{optneurest}) applied with $q=2$ implies that each of these
functions can be approximated from $\mathcal{S}_n$ up to accuracy
$n^{-r/2}$.  Our assumption that $f\in W_{H,r,2}^{\mbox{NN}}$ implies that  each of these constituent functions is Lipschitz
continuous. Hence, it is easy to deduce that, for example, if $P$, $P_1$,
$P_2$ are approximations to the constituent functions $h$, $h_1$,
$h_2$, respectively within an accuracy of $\epsilon$, then
\begin{eqnarray*}
\|h(h_1,h_2)-P(P_1,P_2)\| &\le& \|h(h_1,h_2)-h(P_1,P_2)\|+\|h(P_1,P_2)-P(P_1,P_2)\|\\
& \le& c\left\{\|h_1-P_1\|+\|h_2-P_2\|+\|h-P\|\right\}\le 3c\epsilon,
\end{eqnarray*}
for some constant $c>0$ independent of $\e$. This
leads to (\ref{deepnetapprox}).
$\square$

The constants involved
in $\O$ in (\ref{optneurest}) will depend upon the norms of the
derivatives of $f$ as well as $\sigma$. Thus, when the only a priori
assumption on the target function is about the number of derivatives,
then to \textbf{guarantee} an accuracy of $\epsilon$, we need a
shallow network with $\O(\epsilon^{-q/r})$ trainable parameters. If we assume a hierarchical structure on the target function as in Theorem~\ref{optneurtheo}, then the corresponding deep network yields a guaranteed accuracy of $\epsilon$ only with $\O(\epsilon^{-2/r})$ trainable parameters.

Is this the best? To investigate this question,   we digress and recall the notion of non--linear widths \cite{devore1989optimal}. If $\XX$ is a normed linear space, $W\subset \XX$ be compact,   $M_n :W\to \mathbb R^n$
be a continuous mapping (parameter selection), and $A_n :\mathbb R^n\to
\XX$ be any mapping (recovery algorithm). Then an approximation to
$f$ is given by $A_n(M_n(f))$, where the continuity of $M_n$ means
that the selection of parameters is robust with respect to
perturbations in $f$. The
 nonlinear $n$--width of the compact set $W$ is defined by
\be\label{nwidthdef}
d_n(W)=\inf_{M_n, A_n}\sup_{f\in W}\| f, A_n(M_n(f))\|_\XX.
\ee
We note that the $n$--width depends only on the compact set $W$ and the space $\XX$, and represents the best
that can be achieved by \textbf{any} continuous parameter selection
and recovery processes.  It is shown in   \cite{devore1989optimal} that
$d_n(W_{r,q}^{\mbox{NN}})\ge cn^{-r/q}$ for some constant
$c>0$ depending only on $q$ and $r$. So, the estimate implied by
(\ref{optneurest}) is {\it the best possible} among \textbf{all} reasonable
methods of approximating arbitrary functions in $W_{r,q}^{\mbox{NN}}$, although by
itself, the estimate (\ref{optneurest}) is blind to the process by
which the approximation is accomplished; in particular, this process
is not required to be robust. Similar considerations apply to the estimate (\ref{deepnetapprox}).

\bhag{Shallow networks}\label{shallowsect}
In this section, we announce our results in the context of shallow networks in two settings. One is the setting of neural networks using the ReLU function $x\mapsto |x|\corr{=x_++(-x)_+}$ (Sub-section~\ref{relusect}), and the other is the setting of Gaussian networks using an activation function of the form $\x\mapsto\exp(-|\x-\w|^2)$ (Sub-section~\ref{gausssect}).  It is our objective to generalize these results to the case of deep networks in Section~\ref{deepsect}. 

\corr{Before starting with the mathematical details, we would like to make some remarks regarding the results in this section and in Section~\ref{deepsect}.
\begin{enumerate}
\item It seems unnatural to restrict the range of the constituent functions. Therefore, we are interested in approximating functions on the entire Euclidean space.
\item If one is interested only in error estimates analogous to those in Theorem~\ref{optneurtheo},  then our results need to be applied to functions supported on the unit cube. 
One way to ensure that  that the smoothness is preserved is to consider a smooth extension of the function on the unit cube to the Euclidean space \cite[Chapter~VI]{stein2016singular}, and then multiply this extension by a $C^\infty$ function supported on $[-2,2]^q$, equal to $1$ on the unit cube. However, this destroys the constructive nature of our theorems.
\item A problem of central importance in approximation theory is to determine what constitutes the right smoothness and the right measurement of complexity. The number of parameters or the number of non--linear units is not necessarily the right measurement for complexity. Likewise, the number of derivatives is not necessarily the right measure for smoothness for every approximation process. In this paper, we illustrate this by showing that different smoothness classes and notions of complexity lead to satisfactory approximation theorems.
\end{enumerate}}

\subsection{ReLU networks}\label{relusect}
In this section, we are interested in approximating functions on $\RR^q$ by networks of the form $\x\mapsto\sum_{k=1}^n a_k|\x\cdot\v_k+b_k|$, $a_k, b_k\in\RR$, $\x, \v_k\in\RR^q$. 
The set of all such functions will be denoted by $\mathcal{R}_{n,q}$.  
Obviously, these networks are not bounded  on the whole Euclidean space. Therefore, we will study the approximation in  weighted  spaces, where the norm is defined by
$$
\|f\|_{w,q} =\esssup_{\x\in\RR^q} \frac{|f(\x)|}{\sqrt{|\x|^2+1}}.
$$
The symbol $X_{w,q}$ will denote the set of all continuous functions $f :\RR^q\to \RR$ for which $(|\x|^2+1)^{-1/2}f(\x)\to 0$ as $\x\to\infty$.  We will define a ``differential operator'' $\mathcal{D}$ and smoothness classes $W_{w, \gamma,q}$ in terms of this operator in Section~\ref{relupfsect}. 

In the sequel, we will adopt the following convention. The notation $A\lesssim B$ means $A\le cB$ for some generic positive constant $c$ that may depend upon fixed parameters
in the discussion, such as  $\gamma$, $q$, but independent of the target function and the number of parameters in the approximating network. By $A\sim B$, we
mean $A \lesssim B$ and $B \lesssim A$. 

Our first main theorem is the following Theorem~\ref{relutheo}. 
We note two technical novelties here. One is that the activation function $|\cdot|$ does not satisfy the conditions of Theorem~\ref{optneurtheo}. Second is that the approximation is taking place on the whole Euclidean space rather than on a cube as in Theorem~\ref{optneurtheo}.

\begin{theorem}\label{relutheo}
Let $\gamma>0$,  $n\ge 1$ be an integer, $f\in W_{w,\gamma,q}$. Then there exists $P\in\mathcal{R}_{n,q}$ such that
\be\label{reluapproxbd}
\|f-P\|_{w,q}\lesssim n^{-\gamma/q}\|f\|_{w,\gamma,q}.
\ee
\end{theorem}

\subsection{Gaussian networks}\label{gausssect}
We wish to consider shallow networks where each channel evaluates a
Gaussian non--linearity; i.e., Gaussian networks of the form
\be\label{gaussnetworkdef} 
G(\x)=\sum_{k=1}^n
a_k\exp(-|\x-\x_k|^2),\qquad \x\in\mathbb R^q.  
\ee 
 It is natural to
consider the number of trainable parameters $(q+1)n$ as a measurement
of the complexity of $G$.  However, it is  known  (\cite{convtheo}) that an even more
important quantity that determines the approximation power of
Gaussian networks is the minimal separation among the centers.  For
any subset $\C$ of $\mathbb R^q$, the minimal separation of $\C$ is defined
by 
\be\label{minsepdef} \eta(\C)=\inf_{\x,\y\in\C, \x\not=\y}|\x-\y|.
\ee 
For $n, m>0$, the symbol $\mathcal{N}_{n,m}(\mathbb R^q)$ denotes the
set of all Gaussian networks of the form (\ref {gaussnetworkdef}),
with $\eta(\{\x_1,\cdots,\x_n\})\ge 1/m$.

Let $\XX_q$ be the
space  of continuous functions on $\RR^q$
vanishing at infinity, equipped with the norm
$\|f\|_q=\max_{\x\in\RR^q}|f(\x)|$. 

In order to measure the smoothness of the target function,  we 
need to put conditions not just on the number of derivatives but also
on the rate at which these derivatives tend to $0$ at
infinity. Generalizing an idea from   \cite{freud1972direct,tenswt}, we
 define  first the space $W_{r,q}$ for integer $r\ge 1$ as the set of
all functions $f$ which are $r$ times iterated integrals of functions in $\XX$, satisfying
$$
\|f\|_{r,q}=\|f\|_{q}+\sum_{1\le|\k|_1\le r}\|\exp(-|\cdot|^2)D^\k(\exp(|\cdot|^2)f)\|_{q} <\infty.
$$
 Since one of our goals is to show that our results on the upper
bounds for the accuracy of approximation are the best possible for
individual functions, the class $W_{r,q}$ needs to be refined
somewhat.  Toward that goal, we define next a regularization
expression, known in approximation theory parlance as a
$K$--functional,  by
$$
K_{r,q}(f,\delta)=\inf_{g\in W_{r,q}}\{\|f-g\|_{q}+\delta^r(\|g\|_{q}+\|g\|_{r,q})\}.
$$
We note that the infimum above is over \textbf{all} $g$ in the class
$W_{r, q}$ rather than just the class of all networks. The class
$\mathcal{W}_{\gamma,q}$ of functions which we are interested in is
then defined for $\gamma>0$ as the set of all $f\in \XX_q$ for
which
$$
\|f\|_{\gamma,q}=\|f\|_{q}+\sup_{\delta\in (0,1]}\frac{K_{r,q}(f,\delta)}{\delta^\gamma}<\infty,
$$
for some integer $r\ge \gamma$. It turns out that different choices of
$r$ yield equivalent norms, without changing the class itself. The
following theorem gives a bound on approximation of $f\in \XX_q$
from $\mathcal{N}_{N,m}(\RR^q)$.  
The following theorem is proved in \cite{convtheo}.

\begin{theorem}\label{unidegapptheo}
   Let $\{\C_m\}$ be a sequence of finite subsets
  with $\C_m\subset [-cm,cm]^q$, with 
  \be\label{uniformity}
  1/m \lesssim\max_{\y\in [-cm,cm]^q}\min_{\x\in \C}|\x-\y| \lesssim \eta(\C_m),
  \qquad m=1,2,\cdots.  
  \ee 
Let $1\le p\le\infty$, $\gamma>0$, and $f\in
  \mathcal{W}_{\gamma,q}$. Then for integer $m\ge 1$, there exists
  $G\in \mathcal{N}_{|\C_m|,m}(\RR^q)$ with centers at points in $\C_m$ such
  that 
  \be\label{unidirect} \|f-G\|_{q} \lesssim
  \frac{1}{m^\gamma}\|f\|_{\gamma,q}.  
  \ee 
  Moreover, the coefficients
  of $G$ can be chosen as linear combinations of the data $\{f(\x)
  :\x\in\C_m\}$.
\end{theorem}

We note that the set of centers $\C_m$ can be chosen arbitrarily
subject to the conditions stated in the theorem; \textbf{there is no
  training necessary to determine these parameters}. Therefore, there
are only $\O(m^{2q})$ coefficients to be found by training. This means
that if we assume a priori that $f\in \mathcal{W}_{\gamma,q}$, then
the number of trainable parameters to theoretically guarantee an
accuracy of $\epsilon>0$ is $\O(\epsilon^{-2q/\gamma})$. 
For the unit ball $\mathcal{B}_{\gamma, q}$ of the class $\mathcal{W}_{\gamma,q}$ as defined in Section~\ref{gausssect}, the Bernstein inequality proved in \cite{mohapatrapap} leads to $d_n(\mathcal{B}_{\gamma,q})\sim n^{-\gamma/(2q)}$. 
Thus, the estimate \eref{unidirect} is the best possible in terms of widths. This implies in particular that when the networks are
computed using samples of $f$ to obtain an accuracy of 
$\e$ in the approximation, one needs $\sim \e^{-2q/\gamma}$ samples.
\corr{When $f$ is compactly supported, $\|f\|_{\gamma,q}$ is of the same order of magnitude as the norm of $f$ corresponding to the 
$K$-functional based on the smoothness class $W_r^{\mbox{NN}}$ in Section~\ref{reviewsect}. However, the number of parameters is then not commesurate with the results in that section.} 

We observe that the  width estimate holds for the approximation of the entire class, and hence, an agreement with such width estimate implies only that there exists a possibly pathological function for which the approximation estimate cannot be improved. How good is the estimate in Theorem~\ref{unidegapptheo} for individual functions? If we know that
some oracle can give us Gaussian networks that achieve a given
accuracy with a given complexity, does it necessarily imply that the
target function is smooth as indicated by the above theorems?  The following is a converse to Theorem~\ref{unidegapptheo}
 demonstrating that the accuracy asserted by
these theorems is possible if and only if the target function is in
the smoothness class required in these theorems. It demonstrates also that rather than the number of nonlinearities in the Gaussian network, it is the minimal separation among the centers that
is the ``right'' measurement for the complexity of the networks.
Theorem~\ref{convtheo} below is a refinement of the corresponding result in \cite{convtheo}.

\begin{theorem}
  \label{convtheo}  Let $\{\C_m\}$ be a sequence of finite
  subsets of $\mathbb R^q$, such that for each integer $m\ge 1$,
  $\C_m\subseteq \C_{m+1}$, $|\C_m|\le c\exp(c_1m^2)$, and
  $\eta(\C_m)\ge 1/m$. Further, let $f\in \XX_q$, and for each
  $m\ge 1$, let $G_m$ be a Gaussian network with centers among points
  in $\C_m$, such that 
  \be\label{uniconv_degapprox} 
  \sup_{m\ge
    1}m^\gamma\|f-G_m\|_q <\infty. 
     \ee
  Then $f\in \mathcal{W}_{\gamma,q}$. 

\end{theorem}

\corr{We observe that  Theorem~\ref{unidegapptheo} can be interpreted to give estimates on the degree of approximation by Gaussian networks either in terms of the number of non--linear units, or the number of trainable parameters, or the minimal separation among the centers, or the number of samples of the target function.
Theorem~\ref{convtheo} shows that the right model of complexity among these is the minimal separation among the centers. Using this measurement for complexity yields ``matching'' direct and converse theorems. Based on the results in \cite{eignet}, we expect that a similar theorem should be true also for ReLU networks.}

\bhag{Deep networks}\label{deepsect}
The purpose of this section is to generalize the results in Section~\ref{shallowsect} to the case of deep networks. In Sub-section~\ref{dagsect}, we will formulate the concept of compositional functions in terms of a DAG, and introduce the related mathematical concepts for measuring the degree of approximation and smoothness. The approximation theory results in this context will be described in Sub-section~\ref{deepapproxsect}.

\subsection{General DAG functions}\label{dagsect}

\begin{figure}[h]
\begin{center}
\includegraphics[scale=0.3]{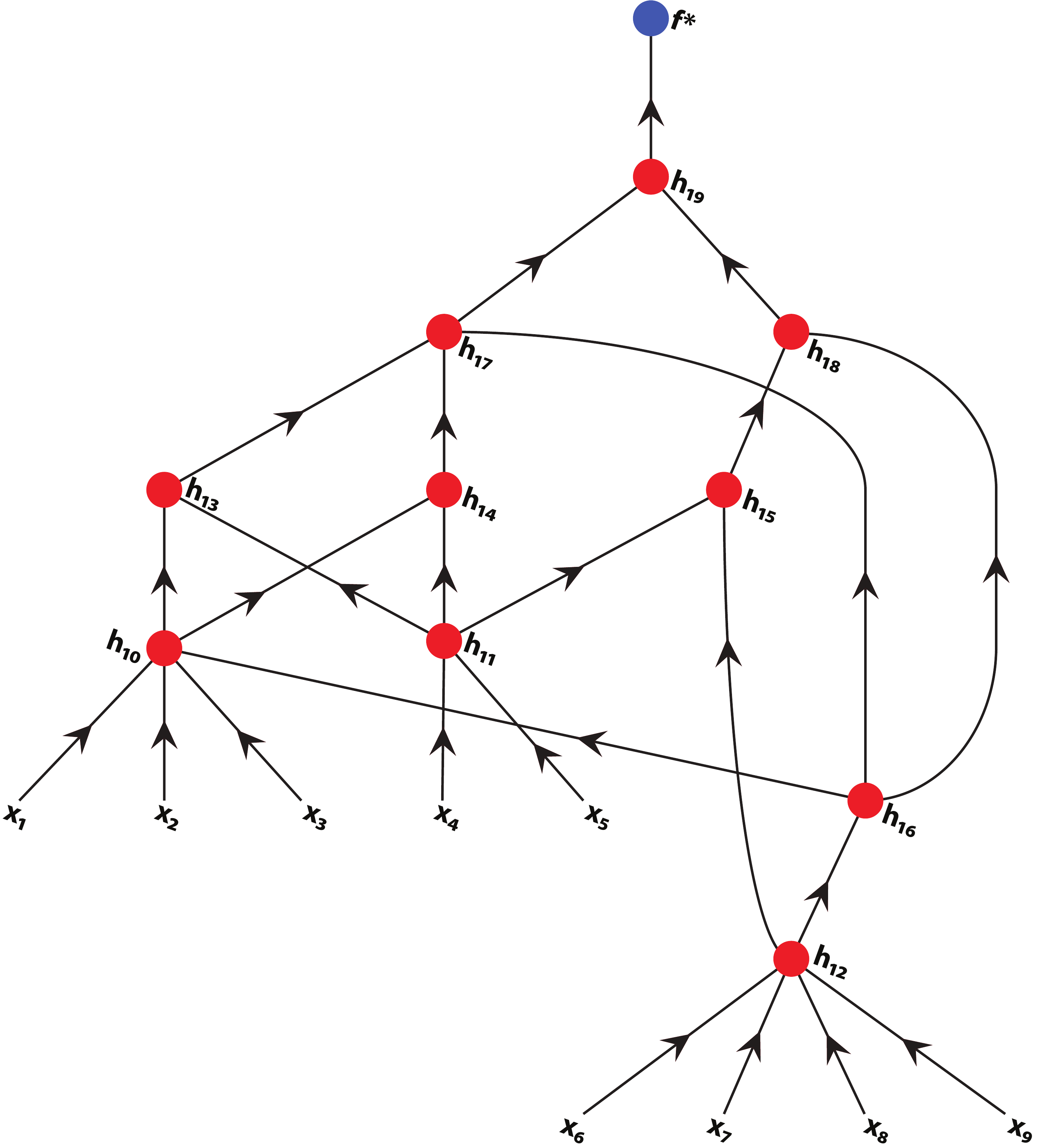}  
\end{center}

\caption{\it An example of a $\mathcal{G}$--function ($f^*$ given in \eref{gfuncexample}). The vertices of the DAG $\mathcal{G}$ are denoted by red dots. The black dots represent the input to the various nodes as indicated by the in--edges of the red nodes, and the blue dot indicates the output value of the $\mathcal{G}$--function, $f^*$ in this example.}
\label{graphpict}
\end{figure}

Let $\mathcal{G}$ be a directed acyclic graph (DAG), with the set of nodes $V$. A $\mathcal{G}$--function is defined as follows. The in-edges to each node of $\mathcal{G}$ represents an input real variable. 
The node itself represents the evaluation of a real valued function of the inputs. The out-edges fan out the result of this evaluation. Each of the source node obtains an input from some Euclidean space. Other nodes can also obtain such an input. We assume that there is only one sink node, whose output is the $\mathcal{G}$-function. For example, the DAG in Figure~\ref{graphpict} represents the $\mathcal{G}$--function
\bea\label{gfuncexample}
\lefteqn{f^*(x_1,\cdots, x_9)=}\nonumber\\
&&h_{19}(h_{17}(h_{13}(h_{10}(x_1,x_2,x_3, h_{16}(h_{12}(x_6,x_7,x_8,x_9))), h_{11}(x_4,x_5)), h_{14}(h_{10},h_{11}), h_{16}), h_{18}(h_{15}(h_{11},h_{12}),h_{16}))\nonumber\\
\eea

We note that if $q$ is the number of source nodes in $\mathcal{G}$, a $\mathcal{G}$--function is a function on $\RR^q$. Viewed only as a function on $\RR^q$, it is not clear whether two different DAG structures can give rise to the same function. Even if we assume a certain DAG, it
is not clear that the choice of the constituent functions is uniquely determined for a given function on $\RR^q$. For our mathematical analysis, we therefore find it convenient to think of a $\mathcal{G}$--function as a set of functions $f=\{f_v :\RR^{d(v)}\to\RR\}_{v\in V}$, rather than a single 
function on $\RR^q$. The individual functions $f_v$ will be called \textit{constituent functions}.

 We adopt the convention that for any function class $\XX(\RR^d)$, the class $\mathcal{G}\XX$ denotes 
the set of $\mathcal{G}$ functions $f=\{f_v\}_{v\in V}$, where each constituent function $f_v\in \XX(\RR^{d(v)})$. We define 
\be\label{gengfuncnormdef}
\|f\|_{\mathcal{G},\XX}=\sum_{v\in V}\|f_v\|_{\XX(\RR^{d(v)})}.
\ee

\subsection{Approximation using deep  networks}\label{deepapproxsect}

First, we discuss the analogue of Theorem~\ref{relutheo} in Section~\ref{relusect} for deep networks conforming to the DAG $\mathcal{G}$. We define the classes $\mathcal{G}X_w$ and $\mathcal{G}W_{w,\gamma}$ in accordance with the notation introduced in Section~\ref{dagsect}, and denote the norm on 
$\mathcal{G}X_w$ (respectively, $\mathcal{G}W_{w,\gamma}$) by $\|\cdot\|_{\mathcal{G},w}$ (respectively, $\|\cdot\|_{\mathcal{G}, w,\gamma}$). The symbol $\mathcal{G}\mathcal{R}_n$ denotes the family of networks $\{P_v\in \mathcal{R}_{n,d(v)}\}_{v\in V}$.  The analogue of Theorem~\ref{relutheo} is the following.

\begin{theorem}
\label{deeprelutheo}
Let $1\le \gamma\le 2$,  $n\ge 1$ be an integer, $f\in \mathcal{G}W_{w,\gamma}$, $d=\max_{v\in V}d(v)$. Then there exists $P\in\mathcal{G}\mathcal{R}_n$ such that
\be\label{deepreluapproxbd}
\|f-P\|_{\mathcal{G},w}\lesssim n^{-\gamma/d}\|f\|_{\mathcal{G},w,\gamma}.
\ee
\end{theorem}

We observe that if $P=\{P_v\}\in \mathcal{G}\mathcal{R}_n$ the number of trainable parameters in each constituent network $P_v$ is $\O(n)$. Therefore, the total number of trainable parameters in $P$ is $\O(|V|n)$. Equivalently, when the target  function is in  $\mathcal{G}W_{w,\gamma}$, one needs $\O((\e/|V|)^{-d/\gamma})$ units in a deep network to achieve an accuracy of at most $\e$. If one ignores the compositional structure of the target function, \eref{reluapproxbd} shows that one needs $O(\e^{-q/\gamma})$ units in a shallow network.  Thus, a deep network conforming to the structure of the target function yields a substantial improvement over a shallow network if $d\ll q$. 

Next, we discuss deep Gaussian networks. As before, the spaces $\mathcal{G}\XX$ and $\mathcal{G}W_{\gamma}$ are as described in Section~\ref{dagsect}, and denote the corresponding norms $\|\cdot\|_{\mathcal{G}\XX}$ (respectively, $\|\cdot\|_{\mathcal{G}W_{\gamma}}$) by $\|\cdot\|_{\mathcal{G}}$ (respectively, $\|\cdot\|_{\mathcal{G},\gamma}$).

The analogue of Theorem~\ref{unidegapptheo} and Theorem~\ref{convtheo} are parts (a) and (b) respectively of the following Theorem~\ref{deepgausstheo}.

\begin{theorem}
\label{deepgausstheo}
{\rm (a)} For each $v\in V$, let $\{\C_{m,v}\}$ be a sequence of finite subsets
as described in Theorem~\ref{unidegapptheo}. Let $\gamma\ge 1$ and
$f\in\mathcal{G}\mathcal{W}_{\gamma}$. Then for integer $m\ge 1$,
there exists $G\in \mathcal{G}\mathcal{N}_{\max|\C_{m,v}|,m}$
with centers of the constituent network $G_v$ at vertex $v$ at points
in $\C_{m,v}$ such that 
\be\label{deepgfuncdirect}
\|f-G\|_{\mathcal{G}} \lesssim
\frac{1}{m^\gamma}\|f\|_{\mathcal{G},\gamma}.  
\ee Moreover, the
coefficients of each constituent $G_v$ can be chosen as linear
combinations of the data $\{f_v(\x) :\x\in\C_{m,v}\}$.\\
{\rm (b)} For each $v\in V$, let $\{\C_{m,v}\}$ be a sequence of
  finite subsets of $\mathbb R^{d(v)}$, satisfying the conditions as
  described in part (a) above. Let $f\in\mathcal{G}\XX$,
  $\gamma>0$, and $\{G_m\in \mathcal{G}\mathcal{N}_{n,m}$\} be a
  sequence where, for each $v\in V$, the centers of the constitutent
  networks $G_{m,v}$ are among points in $\C_{m,v}$, and such that
  \be\label{gfuncconv_degapprox} \sup_{m\ge
    1}m^\gamma\|f-G_m\|_{\mathcal{G}} <\infty.  \ee Then $f\in
  \mathcal{G}\mathcal{W}_{\gamma}$.
\end{theorem}

\vskip -0.5cm
\bhag{Ideas behind the proofs}\label{ideasect}

\subsection{Theorem~\ref{relutheo}.}\label{relupfsect}
The proof of this theorem has two major steps. One is a reproduction formula (\eref{repkern} below), and the other is  the definition of smoothness. Both are based on ``wrapping'' the target function from $\RR^q$ to a function $\mathcal{S}(f)$ (cf. \eref{rqtosqtrans} below) on the   unit Euclidean  sphere $\SS^q$, defined by
$$
\SS^q=\{\u\in\RR^{q+1} : |\u|=1\}.
$$
A parametrization of the upper hemisphere $\SS^q_+=\{\u\in\SS^q :
u_{q+1}>0\}$ of $\SS^q$ is given by
\be\label{basicparam}
u_j=\frac{x_j}{\sqrt{|\x|^2+1}}, \quad j=1,\cdots, q, 
\quad u_{q+1}=(|\x|^2+1)^{-1/2},\quad \u\in\SS^q_+, \ \x\in  \RR^q,
\ee
with the inverse mapping
\be\label{inverseparam}
x_j=\frac{u_j}{u_{q+1}}, \qquad j=1,\cdots, q, \qquad \u\in\SS^q_+,\  \x\in\RR^q.
\ee
Next, we define an operator $\mathcal{S}$ on $X_{w,q}$ by
\be\label{rqtosqtrans}
\mathcal{S}(f)(\u)= |u_{q+1}|f\left(\frac{u_1}{u_{q+1}}, \cdots, \frac{u_q}{u_{q+1}}\right), \qquad f\in X_{w,q}. 
\ee
We note that if $f\in X_{w,q}$, then $(|\x|^2+1)^{-1/2}f(\x)\to 0$ as $|\x|\to\infty$. Therefore, $\mathcal{S}(f)$ is well defined, and defines an even, continuous function on $\SS^q$, equal to $0$ on the ``equator'' $u_{q+1}=0$.

Next, let $\mu^*$ be the Riemannian volume measure on $\SS^q$, with $\mu^*(\SS^q)=\omega_q$. In this subsection, we denote the dimension of the space of all homogeneous spherical polynomials of degree $\ell$ by $d_\ell$, $\ell=0,1,\cdots$, and the set of orthonormalized spherical harmonics on $\SS^q$ by $\{Y_{\ell,k}\}_{k=1}^{d_\ell}$. 
If $F\in L^1(\SS^q)$, then 
\be\label{fourcoeff}
\hat{F}(\ell, k)=\int_{\SS^q} F(\u)Y_{\ell,k}(\u)d\mu^*(\u).
\ee
We note that if $F$ is an even function, then  $\hat{F}(2\ell+1, k)=0$ for $\ell=0,1,\cdots$. 

Next, we recall the addition formula
\be\label{additionformula}
\sum_{k=1}^{d_\ell}Y_{\ell,k}(\u)\overline{Y_{\ell, k}(\v)}=\omega_{q-1}^{-1}p_\ell(1)p_\ell(\u\cdot\v),
\ee
where $p_\ell$ is the degree $\ell$ ultraspherical polynomial with positive leading coefficient, with the set $\{p_\ell\}$ satisfying
\be\label{orthonormal}
\int_{-1}^1 p_\ell(t)p_j(t)(1-t^2)^{q/2-1}dt =\delta_{j,\ell}, \qquad j, \ell =0, 1,\cdots.
\ee
The function $t\to |t|$ can be expressed in an expansion
\be\label{absseries}
|t|\sim p_0-\sum_{\ell=1}^\infty \frac{\ell-1}{\ell(2\ell-1)(\ell+q/2)}p_{2\ell}(0)p_{2\ell}(t), \qquad t\in [-1,1],
\ee
with the series converging on compact subsets of $(-1,1)$. 

We define the $\phi$--derivative of $F$ formally by
\be\label{formderdef}
\widehat{\mathcal{D}_\phi F}(2\ell,k)=\left\{\begin{array}{ll}
\hat{F}(0,0), & \mbox{if  $\ell=0$},\\[1ex]
\disp -\frac{\ell(2\ell-1)(\ell+q/2)p_{2\ell}(1)}{\omega_{q-1}(\ell-1)p_{2\ell}(0)}\hat{F}(2\ell, k), &\mbox{ if $\ell=1,2,\cdots$,}
\end{array}\right.
\ee
and $\widehat{\mathcal{D}_\phi F}(2\ell+1,k)=0$ otherwise.
Then for an even function $F \in L^1(\SS^q)$ for which $\mathcal{D}_\phi F\in L^1(\SS^q)$, we deduce  the reproducing kernel property:
\be\label{repkern}
F(\u)=\int_{\SS^q}|\u\cdot\v|\mathcal{D}_\phi F(\v)d\mu^*(\v).
\ee
A careful discretization of this formula using polynomial approximations of both the terms in the integrand as in \cite{zfquadpap, eignet} leads to a zonal function network of the form $\u\mapsto\sum_{k=0}^n a_k|\u\cdot\v_k|$, $a_k\in\RR$, $\v_k\in\SS^q$, satisfying
\be\label{relufavard}
\left|F(\u)-\sum_{k=0}^n a_k|\u\cdot\v_k|\right|\lesssim n^{-1/q}\esssup_{\v\in\SS^q}|\mathcal{D}_\phi F(\v)|, \qquad \u\in\SS^q.
\ee

Next, we define formally
\be\label{reluderdef}
\mathcal{D}(f)(\x)=(|\x|^2+1)^{1/2}\mathcal{D}_\phi (\mathcal{S}(f))\left(\frac{x_1}{\sqrt{|\x|^2+1}},\cdots,\frac{x_q}{\sqrt{|\x|^2+1}}, \frac{1}{\sqrt{|\x|^2+1}}\right).
\ee

The estimate \eref{relufavard} now leads easily for all $f\in X_{w,q}$ for which $\mathcal{D}(f)\in X_{w,q}$ to
\be\label{relufavardeuclid}
\left|f(\x)-\sum_{k=1}^n \frac{a_k}{\sqrt{|\x_k|^2+1}}|\x\cdot \x_k+1|\right|\lesssim n^{-1/q}\|\mathcal{D} (f)\|_{w,q},
\ee
where $\x_k$ is defined by $(\x_k)_j=(\v_k)_j/(\v_k)_{q+1}$, $j=1,\cdots,q$.

In order to define the smoothness class $W_{w,\gamma,q}$, we first define the $K$--functional
\be\label{relukfunc}
K_w(f,\delta)=\inf\left\{\|f-g\|_{w,q}+\delta\|\mathcal{D}(g)\|_{w,q}\right\},
\ee
where the infimum is taken over all $g$ for which $\mathcal{D}g\in X_{w,q}$.
Finally, the smoothness class $W_{w,\gamma,q}$ is defined to be the set of all $f\in X_{w,q}$ such that
$$
\|f\|_{w,\gamma,q}=\|f\|_{w,q}+\sup_{0<\delta<1}\frac{K_w(f,\delta)}{\delta^\gamma}\corr{<\infty.}
$$
The estimate \eref{relufavardeuclid} then leads to \eref{reluapproxbd} in a standard manner. 

\corr{We remark here that the unit cube $[-1,1]^q$ is mapped to some compact subset of $\SS^q_+$. However, the operator $\mathcal{D}$ does not have an obvious interpretation in terms of ordinary derivatives on the cube.}

\subsection{Theorems~\ref{unidegapptheo} and \ref{convtheo}.}\label{gausspfsect}

In this section, let $\{\psi_j\}$ denote the sequence of orthonormalized Hermite functions; i.e., \cite[Formulas~(5.5.3), (5.5.1)]{szego}
\be\label{hermitedef}
\psi_j(x)= \frac{(-1)^j}{\pi^{1/4}2^{j/2}\sqrt{j!}}\exp(x^2/2)\left(\frac{d}{dx}\right)^j (\exp(-x^2)), \qquad x\in\RR, \ j=0,1,\cdots.
\ee
The multivariate Hermite functions are defined by
\be\label{multihermitedef}
\psi_\j(\x)=\prod_{\ell=1}\psi_{j_\ell}(x_\ell).
\ee
We note that
\be\label{hermiteortho}
\int_{\RR^q} \psi_\j(\z)\psi_\k(\z)d\z =\delta_{\j,\k},\qquad \j,\k\in \ZZ_+^q.
\ee

Using the Mehler formula \cite[Formula~(6.1.13)]{andrews_askey_roy}, it can be shown that
\be\label{hermite_recovery}
\psi_\j(\y)=\frac{3^{|\j|/2}}{(2\pi)^{q/2}}\int_{\RR^q} \exp(-|\y-\w|^2)\exp(-|\w|^2/3)\psi_\j(2\w/\sqrt{3})d\w.
\ee
We combine the results on function approximation and quadrature formulas developed in \cite{mohapatrapap} to complete the proof of Theorem~\ref{unidegapptheo}.

To prove Theorem~\ref{convtheo}, we modify the ideas in \cite{gaussbern} to obtain a Berstein--type inequality for Gaussian networks of the form
\be\label{gaussbernineq}
\|g\|_{r,q}\lesssim m^r\|g\|_q, \qquad g\in N_{N,m}, \quad N\lesssim \exp(cm^2).
\ee 
The proof of Theorem~\ref{convtheo} then follows standard arguments in approximation theory.

\subsection{Results in Section~\ref{deepapproxsect}. }\label{deeppfsect}

Theorems~\ref{deeprelutheo} and \ref{deepgausstheo}(a) follow from Theorems~\ref{relutheo} and \ref{unidegapptheo} respectively by the ``good error propagation property'' as in the proof of Theorem~\ref{optneurtheo}(b) from Theorem~\ref{optneurtheo}(a). 
Our definitions of the norms for function spaces associated with
deep networks ensure that a bound of the form 
\eref{gfuncconv_degapprox} implies a bound of the form \eref{uniconv_degapprox} for each of the constituent functions. Therefore, Theorem~\ref{convtheo} leads to Theorem~\ref{deepgausstheo}(b).
 
\bhag{Blessed representations}\label{sparssect}

As pointed out in Sections~\ref{reviewsect} and \ref{deepsect}, 
\textit{there are deep networks -- for instance of the convolutional
  type -- that can bypass the curse of dimensionality when learning functions blessed
with compositionality.} 
In this section, we explore possible definitions of blessed function representations that can
be exploited by deep but not by shallow networks to reduce the
complexity of learning. We list three examples, each of a different
type.
\begin{itemize}

\item The main example consists of the compositional functions defined
  in this paper in terms of DAGs (Figure \ref{graphpict}). The
  simplest DAG is a binary tree (see Figure
  \ref{example_binary} ) corresponding to compositional
  functions of the type 
  $$
  f(x_1,
  \cdots, x_8) = h_3(h_{21}(h_{11} (x_1, x_2),  h_{12}(x_3, x_4)),
  h_{22}(h_{13}(x_5, x_6),  h_{14}(x_7, x_8))).
  $$
As explained in previous sections, such compositional functions can be
approximated well by deep networks. In particular, if the function form
above has shift symmetry, it takes the form
$$
f(x_1,
  \cdots, x_8) = h_3(h_2(h_1 (x_1, x_2),  h_1(x_3, x_4)),
  h_2(h_1(x_5, x_6),  h_1(x_7, x_8))).
  $$

that can be approximated well by a Deep Convolutional Network (that is
with ``weight sharing'') but not by a shallow one.  This first example is important because
compositionality seems a common feature of algorithms applied to
signals originating from our physical world, such as
images. Not surprisingly, binary-like tree structures (the term binary-like covers
obvious extensions to two-dimensional inputs such as images) represemt
well the architecture of the most successful DCNN.

\item Consider that the proof of
  Theorem~\ref{optneurtheo} relies upon the fact that when $\sigma$
  satisfies the conditions of that theorem, the algebraic polynomials
  in $q$ variables of (total or coordinatewise) degree $<n$ are in the
  uniform closure of the span of $\O(n^q)$ functions of the form
  $\x\mapsto\sigma(\w\cdot\x+b)$.  The advantage of deep nets is due
  to the fact that polynomials of smaller number of variables lead to
  a nominally high degree polynomial through repeated composition.  As
  a simple example, we consider the polynomial
\begin{eqnarray*}
Q(x_1,x_2,x_3,x_4)=(Q_1(Q_2(x_1,x_2), Q_3(x_3,x_4)))^{1024},
\end{eqnarray*}
where $Q_1$, $Q_2$, $Q_3$ are bivariate polynomials of total degree
$\le 2$. Nominally, $Q$ is a polynomial of total degree $4096$ in $4$
variables, and hence, requires $\binom{4100}{4} \approx
(1.17)*10^{13}$ parameters without any prior knowledge of the
compositional structure. However, the compositional structure implies
that each of these coefficients is a function of only $18$
parameters. In this case, the representation which makes deep networks
approximate the function with a smaller number of parameters than
shallow networks is based on polynomial approximation of functions of
the type $g(g(g()))$.

\item As a different example, we consider a function which is a linear
  combination of $n$ tensor product Chui--Wang spline wavelets
  \cite{chuiwaveletbk}, where each wavelet is a tensor product cubic
  spline. It is shown in \cite{chui1994neural, chui1996limitations}
  that is impossible to implement such a function using a shallow
  neural network with a sigmoidal activation function using $\O(n)$
  neurons, but a deep network with the activation function $(x_+)^2$
  can do so. This case is even less general than the previous one but
  it is interesting because shallow networks are provably unable to
  implement these splines using a fixed number of units. In general,
  this does not avoid the curse of dimensionality, but it shows that
  deep networks provide, unlike shallow networks, local and multi--scale approximation since the
  spline wavelets are compactly supported with shrinking supports. 
\item \corr{Examples of
functions that cannot be represented efficiently by shallow networks
have been given very recently by \cite{Telgarsky2015}. The results  in \cite{eldan2015power} illustrate the power of deep networks compared to shallow ones, similar in spirit to \cite{chui1994neural, chui1996limitations}.}
\end{itemize}

The previous examples show three different kinds of ``sparsity'' that
allow a blessed representation by deep networks with a much smaller
number of parameters than by shallow networks. This state of affairs
motivates the following general definition of {\it relative
  dimension}.  Let $d_n(W)$ be the non--linear $n$-width of a function
class $W$. For the unit ball $\mathcal{B}_{\gamma, q}$ of the class
$\mathcal{W}_{\gamma,q}$ as defined in Section~\ref{gausssect}, the
Bernstein inequality proved in \cite{mohapatrapap} leads to
$d_n(\mathcal{B}_{\gamma,q})\sim n^{-\gamma/(2q)}$. In contrast, for
the unit ball $\mathcal{G}\mathcal{B}_\gamma$ of the class we have
shown that $d_n(\mathcal{G}\mathcal{B}_{\gamma})\le
cn^{-\gamma/(2d)}$, where $d=\max_{v\in V}d(v)$.

Generalizing, let $\mathbb{V}$, $\mathbb{W}$ be compact subsets of a
metric space $\XX$, and $d_n(\mathbb{V})$ (respectively,
$d_n(\mathbb{W})$) be their $n$--widths. We define the
\textit{relative dimension} of $d_n(\mathbb{V})$ with respect to
$d_n(\mathbb{W})$ by 
\be\label{reldimdef} 
D(\mathbb{V},
\mathbb{W})=\limsup_{n\to\infty}\frac{\log d_n(\mathbb{W})}{\log
  d_n(\mathbb{V})}.  
  \ee 
  Thus,
$D(\mathcal{G}\mathcal{B}_{\gamma},\mathcal{B}_{\gamma, q})\le d/q$.
This leads us to say that $\mathbb{V}$ is \textit{parsimonious} with
respect to $\mathbb{W}$ if $D(\mathbb{V}, \mathbb{W})\ll 1$.
 
As we mentioned in previous
papers \cite{poggio2015December, mhaskar_poggio_uai_2016}
this definition, and in fact most of the previous results, can be
specialized to the class of Boolean functions which map the Boolean
cube into reals, yielding a number of known \corr{\cite{Hastad1987}} and new results. This application
will be described in a forthcoming paper.
\corr{
\bhag{Conclusion}\label{conclusionsect}
}
\corr{
A central problem of approximation theory is to determine the correct notions of smoothness classes of target functions and the correct measurement of complexity for the approximation spaces.
This definition is dictated by having ``matching'' direct and converse theorems. 
In this paper, we have demonstrated how different smoothness classes lead to satisfactory results for approximation by ReLU networks and Gaussian networks on the entire Euclidean space. 
Converse theorem is proved for Gaussian networks, and results in \cite{eignet} suggest that a similar statement ought to be true for ReLU networks as well. These results indicate that the correct measurement of network complexity is not necessarily the number of parameters. We have initiated a discussion of notions of sparsity which we hope would add deeper insights into this area.}


\end{document}